%% file: main.tex
\documentclass[lettersize,journal]{IEEEtran}

\usepackage[utf8]{inputenc}
\usepackage[T1]{fontenc}
\usepackage{graphicx}
\usepackage{booktabs}
\usepackage{amsmath,amssymb,amsfonts}
\usepackage{nicefrac}
\usepackage{microtype}
\usepackage[ruled,vlined]{algorithm2e}
\usepackage{multirow}
\usepackage{longtable}
\usepackage{array}
\usepackage{pifont}
\usepackage{textcomp}
\usepackage{stfloats}
\usepackage{cite}
\usepackage[dvipsnames]{xcolor}
\usepackage{url}
\usepackage[hidelinks]{hyperref}

\newcommand{\deltaSup}[1]{\raisebox{-0.35ex}{\scriptsize(#1)}}
\newcommand{\xmark}{\ding{55}}

\definecolor{propbg}{RGB}{255,255,0}
\definecolor{locbg}{RGB}{2,255,255}
\begin{document}

\title{CaptchaAgent: Verifiable Visual Reasoning for CAPTCHA Solving}

\author{Shangquan Sun, Ruoyu Chen, Zhirui Liu, Xiang Liu, Chenghao Qian, Wenqi Ren, and Xiaochun Cao%
\thanks{Shangquan Sun is with Nanyang Technological University, Singapore.}%
\thanks{Ruoyu Chen is with the School of Cyber Security, University of Chinese Academy of Sciences, China.}%
\thanks{Zhirui Liu, Wenqi Ren, and Xiaochun Cao are with the School of Cyber Science and Technology, Shenzhen Campus of Sun Yat-sen University, No. 66, Gongchang Road, Guangming District, Shenzhen, Guangdong 518107, P.R. China.}%
\thanks{Xiang Liu is with Beihang University, Beijing, China.}%
\thanks{Chenghao Qian is with the University of Leeds, Leeds, U.K.}%
\thanks{This work has been submitted to the IEEE for possible publication. Copyright may be transferred without notice, after which this version may no longer be accessible.}}

\maketitle

\input{sec/abst}


\input{sec/intro}
\input{sec/relwork}
\input{sec/method}

\input{sec/implementation_details}
\input{sec/exp}
\input{sec/benchmark_details}

\input{sec/res}
\input{sec/qualitative_details}
\input{sec/security_analysis}
\input{sec/limitations}
\input{sec/concl}

\bibliographystyle{IEEEtran}
\bibliography{main}

\end{document}


\title{Supplementary Material for ``CaptchaAgent: Verifiable Visual Reasoning for CAPTCHA Solving''}
\author{Shangquan Sun, Ruoyu Chen, Zhirui Liu, Xiang Liu, Chenghao Qian, Wenqi Ren, and Xiaochun Cao}
\maketitle

\input{sec/supp}

\bibliographystyle{IEEEtran}
\bibliography{main}

%% file: sec/abst.tex
\begin{abstract}
Modern reasoning CAPTCHAs require fine-grained perception, compositional relation reasoning, and grounded browser interaction. End-to-end vision-language model agents remain brittle in this setting because a missed object, incorrect count, or confused relation can propagate directly to the final action. To address this problem, we present \textbf{CaptchaAgent}, a structured visual reasoning framework for interactive CAPTCHA solving. Instead of predicting an action directly from a screenshot, CaptchaAgent first constructs a visual intermediate representation (V-IR) that explicitly records entities, attributes, and spatial or topological relations. A symbol-only ASCII protocol further requires the parser to commit to a concrete object set and count. Building on this global representation, localized evidence refinement revisits ambiguous regions at higher effective resolution, while counterfactual consistency verification tests whether the resulting action remains stable under task-preserving transformations. Across three proprietary backbones on Open CaptchaWorld, CaptchaAgent improves Pass@1 by 21.3--25.8 percentage points over a commercially available browser agent baseline evaluated with the same backbones. On the separate offline Oedipus-data benchmark, it further improves average Pass@1 by 9.2--12.3 percentage points over the Oedipus solver across three matched backbones. The intermediate traces show that exact counting, continuous geometry, and ordered execution remain major sources of automation difficulty for the evaluated agents, providing practical guidance for designing more resilient compositional CAPTCHAs.
\end{abstract}

%% file: sec/intro.tex
\section{Introduction}
\label{sec:intro}

CAPTCHAs (\emph{Completely Automated Public Turing tests to tell Computers and Humans Apart}) are a long-standing security mechanism against automated abuse and have also been studied as a building block for graphical authentication~\cite{von2003captcha,von2008recaptcha,zhu2014captcha}.
Their security margin has repeatedly narrowed with advances in optical character recognition (OCR), deep visual recognition, and data-driven CAPTCHA solving~\cite{sivakorn2016robot,shi2020text,tang2018research}.
Recent CAPTCHA-style puzzles therefore combine cluttered layouts, object relations, counting, rotation, and interactive actions rather than relying on single-object recognition.
Solving them requires not only recognizing visible content, but also constructing a coherent visual state, reasoning over relations, and converting the result into a valid browser action.
CAPTCHA provides a security-relevant and objectively checked environment in which errors at any of these stages directly determine success or failure.

Current vision-language model (VLM) and graphical user interface (GUI) agents commonly predict actions directly from pixels or interleave action prediction with an unconstrained natural-language chain of thought (CoT)~\cite{zheng2024gpt,qin2025ui,wang2025opencua}.
Recent CAPTCHA solvers likewise combine multimodal prompting with structured action spaces, repeated model calls, or task-specific decomposition~\cite{teoh2025captchas,deng2025oedipus,luo2025open}.
These approaches can describe a scene or revise an action, but they do not generally require an explicit, count-committed list of task-relevant entities and their relations to persist across perception, correction, and execution.
This distinction matters for reasoning CAPTCHAs: one omitted object, wrong count, or confused spatial relation can invalidate the interaction, while an unconstrained CoT need not expose the visual state on which the action depends.
The technical gap is therefore not merely stronger visual recognition, but an explicit intermediate state that supports global relational reasoning, targeted correction of local evidence, and action-level verification before execution.

To address this gap, we propose CaptchaAgent, a structured visual reasoning framework for interactive CAPTCHA solving with three components.
First, it converts the puzzle image into a visual intermediate representation (V-IR) that enumerates entities, attributes, and spatial or topological relations.
The V-IR is expressed as a symbol-only ASCII box diagram whose legend commits to an exact object inventory and makes missing or inconsistent relations visible.
Second, localized evidence refinement re-observes high-resolution regions around candidate actions without replacing the global inventory.
Third, counterfactual consistency verification checks action signatures under task-preserving transformations such as color permutations, shifts, and instruction-aware flips.
We operationalize verification as signature-level consistency under these transformations.
Together, these components retain a structured scene state from perception through grounded browser execution.
As illustrated in Fig.~\ref{fig:agent_comparison}, CaptchaAgent differs from direct and CoT-based agents by placing an explicit V-IR between page observation and action prediction, rather than relying on an implicit visual state or a free-form textual rationale.

\input{sec_fig/agent_comparison}

Across 20 Open CaptchaWorld task families, CaptchaAgent improves Pass@1 by 21.3--25.8 percentage points over BU-Agent, a commercially available browser automation agent adopted as the benchmark baseline~\cite{muller2024browseruse,luo2025open}, when both systems use the same proprietary VLM backbone.
On the separate offline Oedipus-data benchmark, CaptchaAgent also improves average Pass@1 by 9.2--12.3 percentage points over the reported Oedipus solver across three matched backbones.
Component ablations show complementary gains from explicit representation, local refinement, and consistency checking.
The resulting traces reveal a security-relevant pattern: discrete relational and interaction-heavy puzzles are more amenable to structured solving, whereas exact counting, continuous geometry, and ordered execution remain persistent bottlenecks. These observations identify which challenge properties current agents can externalize and which continue to preserve automation difficulty. Behavioral web-bot detection, webpage-level security analysis, and rate limits remain complementary defense layers~\cite{cabri2022quantum,purwanto2022phishsim}.

In summary, this work makes the following contributions:
\begin{enumerate}
    \item We propose a structured visual reasoning framework that converts CAPTCHA scenes into an explicit entity--relation representation before grounded action prediction.
    \item We introduce an exact-count-committed ASCII V-IR together with localized evidence refinement and task-preserving counterfactual verification, enabling global--local reasoning and action-level consistency checking.
    \item We demonstrate consistent improvements across three proprietary VLM backbones on 20 interactive CAPTCHA families and three backbones on Oedipus-data. Component ablations and intermediate traces further identify complementary method gains and security-relevant failure modes.
    \item We identify a practical distinction among CAPTCHA families: discrete entity--relation structures are increasingly amenable to automation, while metric geometry, exact counting, and ordered interaction remain more resistant for the evaluated agents, offering guidance for challenge design.
\end{enumerate}

%% file: sec_fig/agent_comparison.tex
\begin{figure}[t]
  \centering
  \includegraphics[width=\columnwidth]{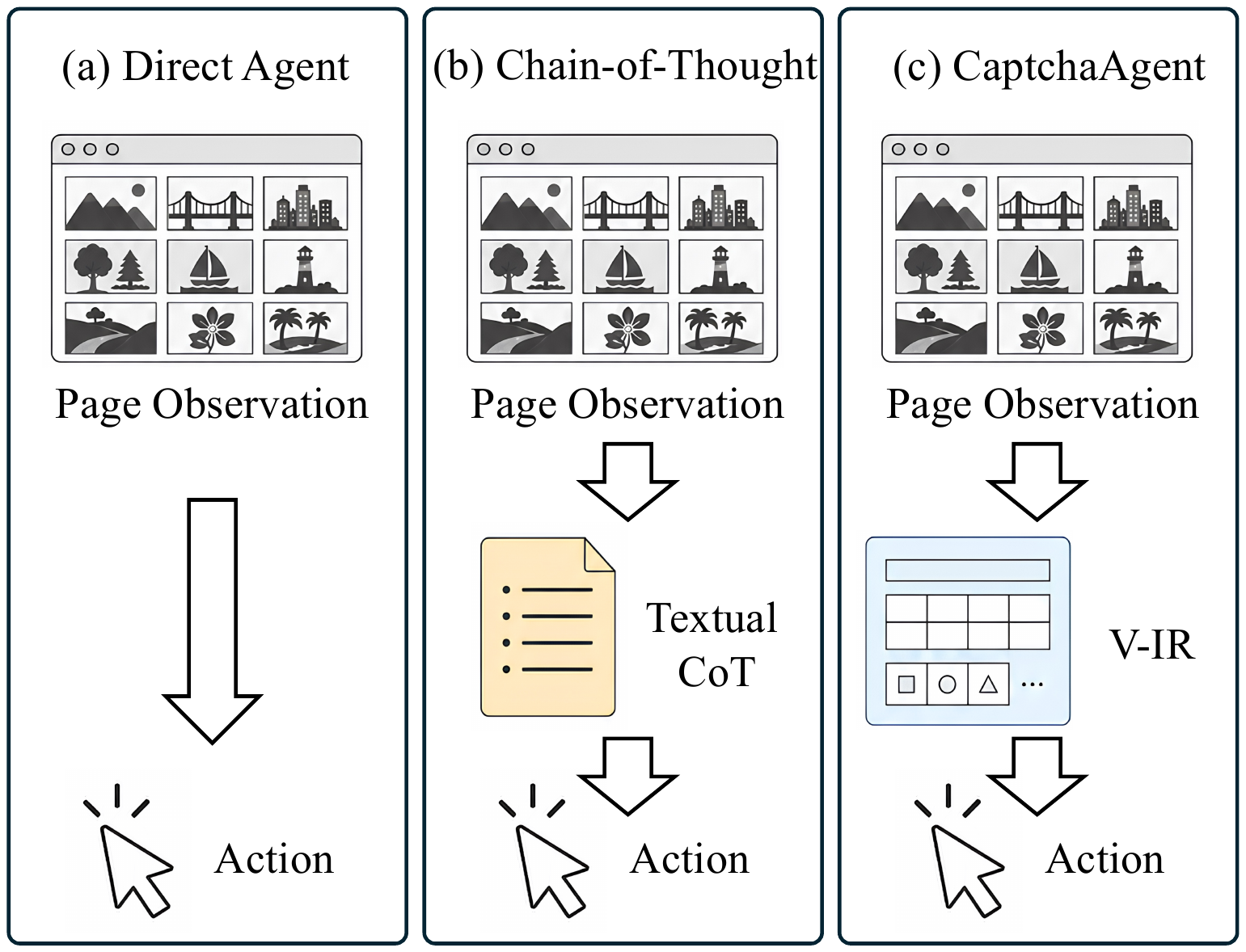}
  \caption{Comparison of action-generation paradigms. Direct agents map observations to actions, chain-of-thought (CoT) agents introduce a free-form textual rationale, and CaptchaAgent instead constructs an explicit V-IR before action prediction. Unlike direct prediction or free-form CoT, the V-IR commits the agent to an explicit object set and relation structure that can be checked before execution.}
  \label{fig:agent_comparison}
\end{figure}

%% file: sec/relwork.tex
\section{Related Work}
\label{sec:related_works}

\paragraph{Multimodal LLMs}
Multimodal large language models (MLLMs) extend LLMs with visual encoders to ground generation in images, enabling unified perception-and-language reasoning. 
Early work such as Flamingo~\cite{alayrac2022flamingo} demonstrated few-shot multimodal prompting via cross-attention over frozen vision features, while later lines improved alignment and instruction following through pretraining and multimodal instruction tuning, e.g., BLIP-style architectures~\cite{li2023blip} and LLaVA-style visual instruction models~\cite{liu2023visual}. 
Open-source systems like MiniGPT-4~\cite{zhu2023minigpt}, InternVL~\cite{chen2024internvl}, and Qwen-VL~\cite{bai2023qwen} further explored scaling, data mixtures, and stronger visual-language alignment to enhance compositional understanding and generalization. 
Despite rapid progress, MLLMs still struggle with fine-grained spatial relations, multi-step visual reasoning, and reliability under distribution shifts, motivating structured consistency checks beyond purely implicit textual inference.
\paragraph{Agents for interactive environments and tool use}
A growing body of work studies LLM/MLLM-based agents that plan, act, and adapt in interactive settings, emphasizing closed-loop decision making over static one-shot prediction~\cite{yao2022react,wang2023voyager}. 
Many agent frameworks decompose behavior into iterative reasoning and action selection, maintain intermediate state, and revise hypotheses as new observations arrive~\cite{ouyang2023autoplan,sun2023adaplanner,rozanov2025stateact}. 
To extend capability and reliability, agents frequently integrate external tools (e.g., search, code execution, structured APIs) and learn to invoke them effectively as part of the policy~\cite{gou2023tora,feng2025retool,qian2025toolrl,wei2022chain}. 
Recent directions further explore training and optimization of agentic policies for long-horizon tasks, including reinforcement learning or preference-based improvements for tool use and planning~\cite{li2025torl,zhang2025nemotron}. 
However, interactive visual challenges like CAPTCHAs remain underexplored in agent research: they demand tight coupling between perception and action, iterative evidence gathering, and robust stateful reasoning under ambiguity, which are not directly addressed by most general-purpose agent frameworks.
\paragraph{GUI agents} 
Advances in vision-language modeling have also accelerated research on GUI agents that map pixel observations to low-level actions (click, scroll, type) across web and mobile interfaces~\cite{nguyen2025gui,zheng2024gpt}. 
Corresponding benchmarks such as Mind2Web~\cite{deng2023mind2web}, ScreenSpot~\cite{cheng2024seeclick}, and Android Control~\cite{li2024effects} emphasize open-ended interaction and generalization across platforms. 
On the modeling side, recent systems improve fine-grained UI perception via high-resolution encoders and specialized visual backbones, e.g., CogAgent~\cite{hong2024cogagent} and Ferret-UI~\cite{you2024ferret}. 
In parallel, structured reasoning has become increasingly standard in GUI agents: approaches like UI-TARS~\cite{qin2025ui} and OpenCUA~\cite{wang2025opencua} interleave perception with explicit reasoning steps, and open-weight multimodal foundation models increasingly support direct action generation and tool invocation~\cite{bai2025qwen3,bai2023qwen}. 
These trends suggest GUI agents are approaching general-purpose interaction, yet their robustness on adversarial, human-gated interfaces remains under-characterized.
\paragraph{Structured visual reasoning}
Recent MLLMs have improved at grounded instruction following and multi-step image reasoning, driven by stronger vision encoders and large-scale multimodal instruction tuning~\cite{alayrac2022flamingo,liu2023visual,li2023blip,zhu2023minigpt,chen2024internvl,bai2023qwen}.
Benchmarks increasingly target visual reasoning such as math and diagram understanding, compositional QA, and multimodal knowledge~\cite{lu2023mathvista,yue2024mmmu,liu2024mmbench}.
Recent work on spatial mental modeling also shows that explicit internal maps can improve spatial reasoning from limited views~\cite{yin2025spatialmentalmodeling}.
The security of structured visual states is not automatic: adversarial attacks on scene graph generation can corrupt object and relation predictions even when the representation itself is explicit~\cite{zhao2024adversarial}.
CaptchaAgent adapts this structured-representation direction to small, interactive CAPTCHA scenes through an inspectable ASCII V-IR and externally checked browser actions.
This setting makes it possible to trace representation errors into browser actions.
\paragraph{Test-time consistency}
Self-consistency and test-time augmentation are widely used to reduce variance by sampling multiple reasoning paths or transformed inputs and aggregating their outputs~\cite{wang2022selfconsistency}.
Related security work studies robustness under physical transformations, localized adversarial patches, and learned separation of natural images from adversarial perturbations~\cite{feng2023robust,li2024physical,qing2024detection}.
Our counterfactual consistency check instantiates this idea at the action level by applying task-preserving image and instruction transformations and aggregating the resulting action signatures.
The key implementation constraint is semantic validity; transformations are disabled when they would change the target rule, such as rotations in rotation-matching puzzles.
\paragraph{Evolution of CAPTCHAs} 
CAPTCHAs began as distorted text recognition tasks designed around the then-large gap between human reading and OCR (Optical Character Recognition)~\cite{von2003captcha}. 
As OCR and deep learning improved, CAPTCHA design shifted from ``hard-to-read'' to ``hard-to-reason'' problems~\cite{singh2014survey,hitaj2020capture,tang2018research}. 
Adversarial-example-based CAPTCHA generation represents another attempt to restore this capability gap against learned solvers~\cite{osadchy2017no}. Modern CAPTCHAs increasingly incorporate visual-semantic components such as image classification, object localization, and spatial reasoning~\cite{searles2023empirical}, and many introduce multi-step interaction and behavioral signals to distinguish humans from scripts. 
Reasoning-oriented CAPTCHAs further raise the bar by requiring compositional logic and contextual understanding~\cite{gossweiler2009s,gao2021research}, which makes them a natural stress test for perception-reasoning-action systems.
\paragraph{Automated CAPTCHA solving}
As CAPTCHAs have grown more complex, automated solvers have also advanced. Earlier pipelines typically focused on narrow CAPTCHA families with task-specific detectors or classifiers, performing well when test distributions matched their design (e.g., GeeSolver~\cite{zhao2023geesolver} and Plesner et al.~\cite{plesner2024breaking}). More recent VLM-based frameworks aim for greater generality, including PhishDecloaker~\cite{teoh2024phishdecloaker}, Halligan~\cite{teoh2025captchas}, and Oedipus~\cite{deng2025oedipus}, often using structured action spaces or domain-specific decompositions. 
However, many general solvers still rely on repeated model calls, heuristics, or explicit search loops, increasing latency and reducing interaction coherence, and it remains unclear whether modern GUI agents can solve CAPTCHAs as a native capability rather than via bespoke pipelines.

\paragraph{Positioning of CaptchaAgent}
Specialized CV solvers use task-specific training and parsing for one known challenge family, while VLM CAPTCHA solvers seek transfer across several types and GUI agents additionally convert perception into browser actions. CaptchaAgent spans the latter two settings. Its distinguishing design combines an explicit count-committed entity--relation representation, targeted local correction, and task-preserving action-consistency checks. The evaluation measures how this structured controller affects interactive solving across multiple CAPTCHA families, and the resulting stage traces support analysis of the remaining failure modes.

%% file: sec/method.tex
\section{Method}
Figure~\ref{fig:framework} gives an overview of CaptchaAgent. At each browser interaction step, the system isolates the puzzle region and object proposals, constructs a structured visual state, revisits local evidence around a candidate action, and checks the resulting action under a bounded set of task-preserving transformations. This section first defines webpage observation and executable grounding, then presents the V-IR, localized evidence refinement, counterfactual consistency verification, and execution checks that form the controller.

\input{sec_fig/arch}

\subsection{Webpage Processing}
CaptchaAgent solves each puzzle as a short-horizon web interaction. At browser interaction step $t$, it observes
\begin{equation}
o_t=\{I_t,x_t,e_t\}, \qquad e_t=\{e_t^{(i)}\}_{i=1}^{M_t},
\end{equation}
where $I_t$ is the screenshot, $x_t$ is the visible page text, and $e_t$ is an indexed Document Object Model (DOM) snapshot containing $M_t$ elements.
Most challenges are centered on a rendered image. Let $\mathcal{D}_t$ denote the set of image-bearing DOM indices. We select the largest such element as the primary puzzle image:
\begin{equation}
i_t^{\mathrm{img}}=\arg\max_{i\in\mathcal{D}_t}\mathrm{area}(e_t^{(i)}), \qquad
J_t=\mathrm{render}(e_t^{(i_t^{\mathrm{img}})}).
\end{equation}
Here, $J_t$ is the rendered crop of the primary image-bearing element. Operating on $J_t$ removes unrelated browser content while preserving a known mapping from local image coordinates to the underlying DOM element, so visually inferred points can be executed in the browser.

Within $J_t$, we use Grounding-DINO~\cite{liu2024grounding}, an open-vocabulary object detector, to propose object-level regions:
\begin{equation}
\{(b_t^{(j)},\ell_t^{(j)})\}_{j=1}^{K_t}=\mathrm{GDINO}(J_t), \qquad
Z_t^{(j)}=\mathrm{crop}(J_t;b_t^{(j)}),
\end{equation}
where $b_t^{(j)}$, $\ell_t^{(j)}$, and $Z_t^{(j)}$ are the bounding box, detector label, and crop for proposal $j$. These proposals form the patch set $\mathcal{P}_t=\{(j,b_t^{(j)},Z_t^{(j)})\}_{j=1}^{K_t}$ used for object-level references and local inspection.

\noindent\textit{Action representation.}
For the evaluated Open CaptchaWorld tasks, the executable action family is $\texttt{action}_t\in\{\textsc{Click},\textsc{Mark},\textsc{Drag},\textsc{Hold}\}$. An executable plan has the form
\begin{equation}
\pi_t=(\texttt{action}_t,i_t,(r_{x,t},r_{y,t}),(d_{x,t},d_{y,t}),T_{\mathrm{hold},t}),
\label{eq:action_plan}
\end{equation}
where $i_t$ is the target DOM index and $(r_{x,t},r_{y,t})\in[0,1]^2$ is an element-local start or click point. The optional $(d_{x,t},d_{y,t})$ specifies the horizontal and vertical drag offset in Cascading Style Sheets (CSS) pixels, so the drag endpoint is the start point plus this offset. The optional $T_{\mathrm{hold},t}$ specifies the hold duration in milliseconds. For set-selection tasks, \textsc{Mark} is implemented as one or more grounded clicks on selectable elements. In contrast, reasoning-only tools do not enter the executable action set.

When the plan refers to detector proposal $j$, the center of $b_t^{(j)}$ is converted from pixels in $J_t$ to normalized coordinates within the primary image element $e_t^{(i_t^{\mathrm{img}})}$. The target DOM index is then $i_t^{\mathrm{img}}$, and the normalized point supplies $(r_{x,t},r_{y,t})$. This conversion connects object-level visual references to executable browser coordinates.

\subsection{Captcha Solving}
CaptchaAgent computes one plan, executes it, and re-observes the page because many puzzles refresh or reveal new state after each action. The complete controller is denoted by $\Phi$:
\begin{equation}
\hat{\pi}_t=\Phi(J_t,x_t,e_t,\mathcal{P}_t;m_{t-1}), \qquad
o_{t+1}\sim \mathcal{E}(o_t,\hat{\pi}_t),
\end{equation}
where $\hat{\pi}_t$ is the final executable plan, $m_{t-1}$ is compact puzzle-scoped memory, and $\mathcal{E}$ is the browser/environment transition induced by DOM updates, rendering latency, and the challenge backend. The following modules define $\Phi$.

\input{sec_fig/ascii_example}

\subsubsection{Visual Intermediate Representation}
\label{sec:vir}
\label{sec:ascii}
A frozen VLM, used as the global scene parser, produces
\begin{equation}
(\mathcal{V}_t,m_t^{(0)})=f_{\theta}(J_t,x_t,\mathcal{P}_t,m_{t-1}),
\end{equation}
where $\theta$ denotes the fixed VLM parameters, $\mathcal{V}_t=(\mathrm{grid}_t,\mathrm{legend}_t)$ is the ASCII diagram and legend, and $m_t^{(0)}$ is a compact memory update. The protocol imposes three constraints: (1) commit to an exact entity count $N_t$; (2) place only markers and spatial lines inside the box while recording attributes in the legend; and (3) state special relations for entity pairs, including containment, occlusion, contact, and links, or explicitly mark them absent.

Figures~\ref{fig:text_reasoning} and~\ref{fig:ascii_vir} present two parallel cases, \textsc{Click Order} and \textsc{Place Dot}, respectively. Each case juxtaposes the web GUI, a text-only CoT, and the proposed ASCII-box V-IR. The CoT can mention object properties and approximate locations, but it need not commit to a complete scene state. By comparison, the V-IR separates spatial markup from semantic attributes, fixes the object inventory, and records relations in a repeatable format. This structure makes an omitted or duplicated entity easier to locate in the intermediate trace and gives subsequent local refinement and transformed-view comparison a shared visual state.

\subsubsection{Plan Inference}
\label{sec:plan}
After the V-IR fixes the scene state, an action decoder maps it to the executable representation in~\eqref{eq:action_plan}:
\begin{equation}
\pi_t^{(0)}=\mathrm{Dec}(\mathcal{V}_t,x_t,e_t,\mathcal{P}_t).
\end{equation}
Here, $\pi_t^{(0)}$ is the initial plan and $\mathrm{Dec}$ is the VLM action-decoding step constrained to the four action families and argument fields defined above. A plan may target a DOM element directly or select a detector proposal that is converted to browser coordinates as described in Section~III-A.

\subsubsection{Localized Evidence Refinement}
\label{sec:localrefine}
The global V-IR establishes object inventory and relations, but small symbols, thin connections, and subtle rotations may still be unresolved. Given the candidate interaction point $\mathbf{u}_t$ in $J_t$, localized evidence refinement extracts
\begin{equation}
R_t=\mathrm{crop}(J_t;\mathrm{window}(\mathbf{u}_t,s))
\end{equation}
and re-parses the region at higher effective resolution:
\begin{equation}
(\pi_t^{r},m_t^{r})=\textsc{Refine}(\mathcal{V}_t,\pi_t^{(0)},R_t,x_t,\mathcal{P}_t,m_t^{(0)};\theta).
\end{equation}
The refined plan $\pi_t^{r}$ changes the initial plan only when local evidence supports a correction; it does not replace the global entity inventory. This global-to-local transition is especially useful for small text, icon matching, and occlusion cues.

\subsubsection{Counterfactual Consistency Verification}
\label{sec:cf}
CaptchaAgent next evaluates the refined decision under a pre-defined, task-specific set $\mathcal{T}$ of deterministic, semantics-preserving transformations. The set includes spatial shifts, flips accompanied by left/right instruction rewrites, and color-channel permutations accompanied by color-token rewrites. For $\tau\in\mathcal{T}$,
\begin{equation}
(J_t^{(\tau)},x_t^{(\tau)})=(\tau(J_t),\rho_\tau(x_t)),
\end{equation}
where $\rho_\tau$ rewrites instruction tokens so that the transformed task remains equivalent. Patch boxes are transformed geometrically with the image. The transformed puzzle is parsed and decoded into $\pi_t^{(\tau)}$, which is then mapped back to the original coordinate frame.

Direct equality between raw coordinates is too sensitive for plan comparison. We therefore define the plan-signature mapping $\sigma$ as
\begin{equation}
\sigma(\pi)=\big(\texttt{action},i,\mathrm{bin}(r_x,r_y),
\mathrm{bin}(d_x,d_y),\mathrm{bin}(T_{\mathrm{hold}})\big),
\end{equation}
where $\mathrm{bin}(\cdot)$ coarsely quantizes continuous action parameters. CaptchaAgent first evaluates a small transformation subset $\mathcal{T}_1$ and forms the signature multiset
\begin{equation}
\mathcal{S}_t^{(1)}=[\sigma(\pi_t^{r})]
\uplus[\sigma(\pi_t^{(\tau)}):\tau\in\mathcal{T}_1].
\end{equation}
Here, $\uplus$ denotes multiset union, which preserves repeated signatures for majority voting. If these signatures disagree, a second bounded subset $\mathcal{T}_2$ is added. The selected signature is
\begin{equation}
\hat{s}_t=\mathrm{mode}\!\left(\mathcal{S}_t^{(1)}\uplus
[\sigma(\pi_t^{(\tau)}):\tau\in\mathcal{T}_2]\right).
\end{equation}
The final executable plan $\hat{\pi}_t$ is a plan in the evaluated set whose signature equals $\hat{s}_t$, mapped to the original frame. Persistent ties are treated as uncertainty rather than model confidence. Transformations that alter the task semantics are excluded; for example, rotation is not used for rotation-matching puzzles.

\subsubsection{Execution Robustness}
Before executing $\hat{\pi}_t$, CaptchaAgent enforces the remaining interaction budget and verifies that the target DOM element is still rendered. For target $i_t$, the validity check is
\begin{equation}
\mathrm{Valid}(i_t)=\mathbb{I}[\mathrm{quads}(e_t^{(i_t)})\neq\emptyset],
\end{equation}
where $\mathrm{quads}(e_t^{(i_t)})$ denotes the browser layout quadrilaterals associated with the rendered element. Invalid or stale targets trigger a bounded wait or re-parse. The memory $m_t$ records instruction changes and previously executed patch or DOM indices needed for multi-step continuity.

%% file: sec_fig/arch.tex
\begin{figure*}[t]
  \centering
  \includegraphics[width=0.98\textwidth]{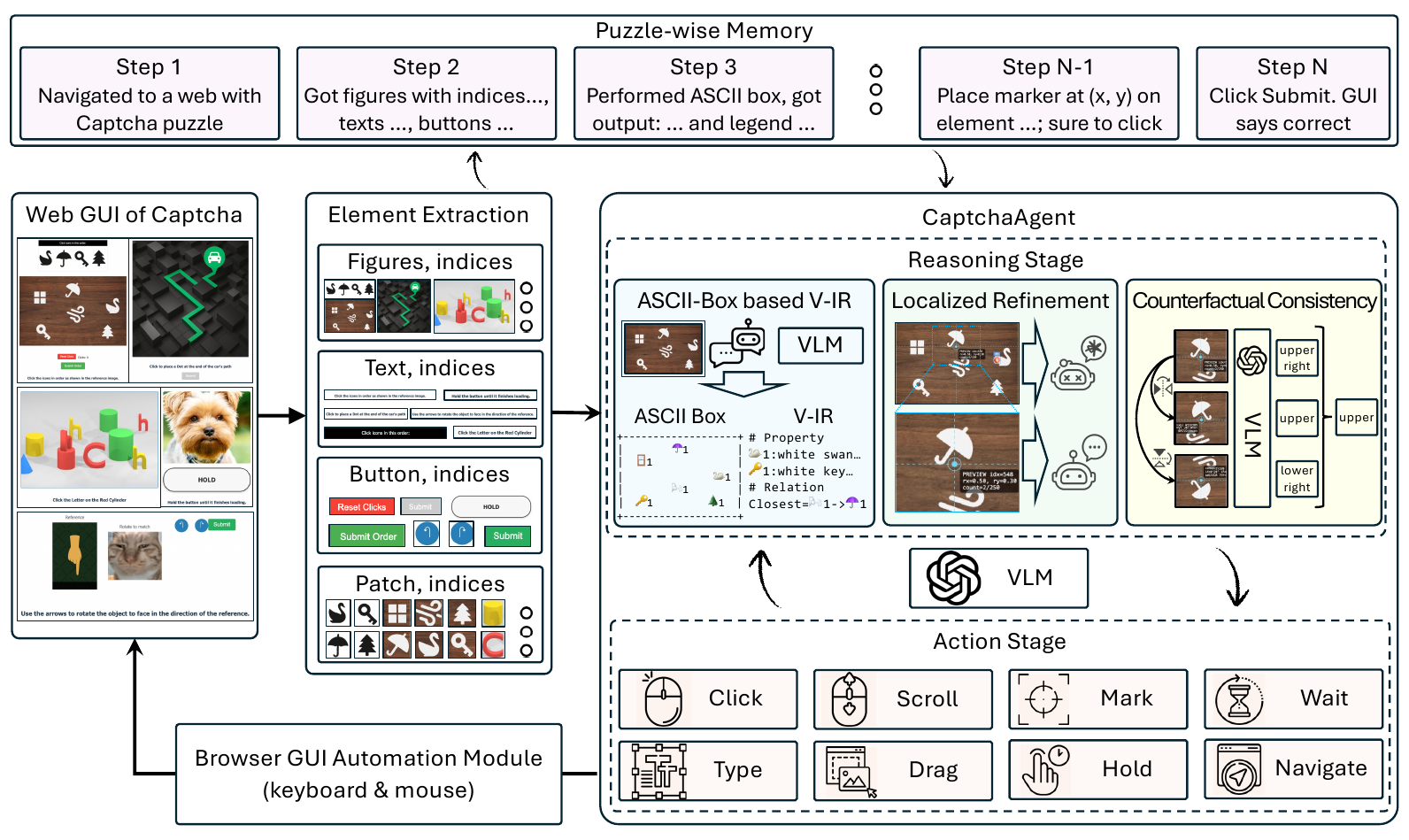}
\caption{Overall framework of CaptchaAgent. At each step, the agent extracts the puzzle image, constructs a V-IR, refines ambiguous local evidence, and applies bounded consistency voting before selecting an action.}
  \label{fig:framework}
\end{figure*}

%% file: sec_fig/ascii_example.tex
\begin{figure}[t]
  \centering
  \includegraphics[page=1,width=\columnwidth]{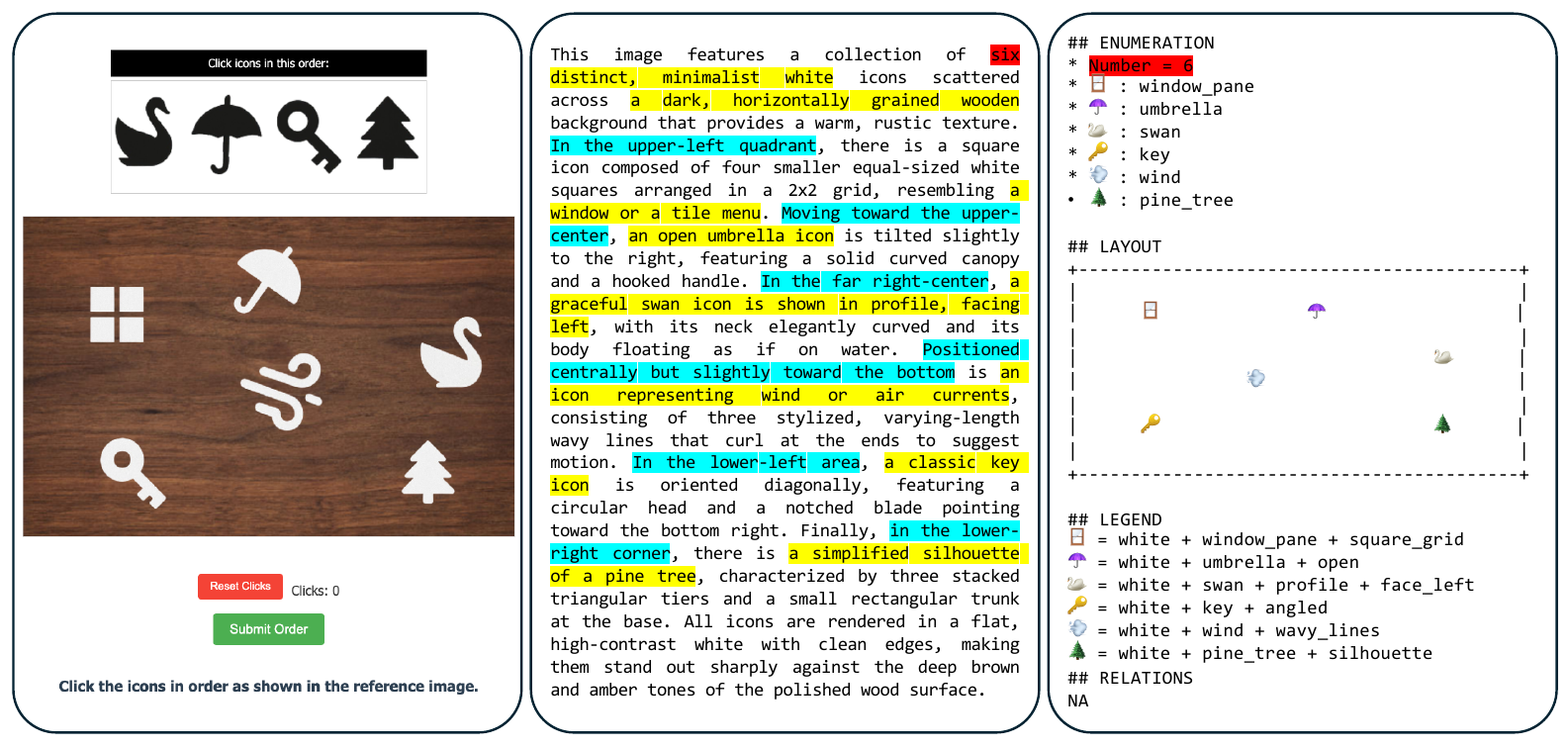}
  \caption{Visual reasoning for a \textsc{Click Order} CAPTCHA. From left to right: the web GUI, a text-only CoT that captures \colorbox{propbg}{object attributes} and \colorbox{locbg}{coarse object locations} but may remain verbose or miss fine-grained spatial cues, and the proposed ASCII-box V-IR that retains the object count, spatial layout, key attributes, and relations.}
  \label{fig:text_reasoning}
\end{figure}

\begin{figure}[t]
  \centering
  \includegraphics[width=\columnwidth]{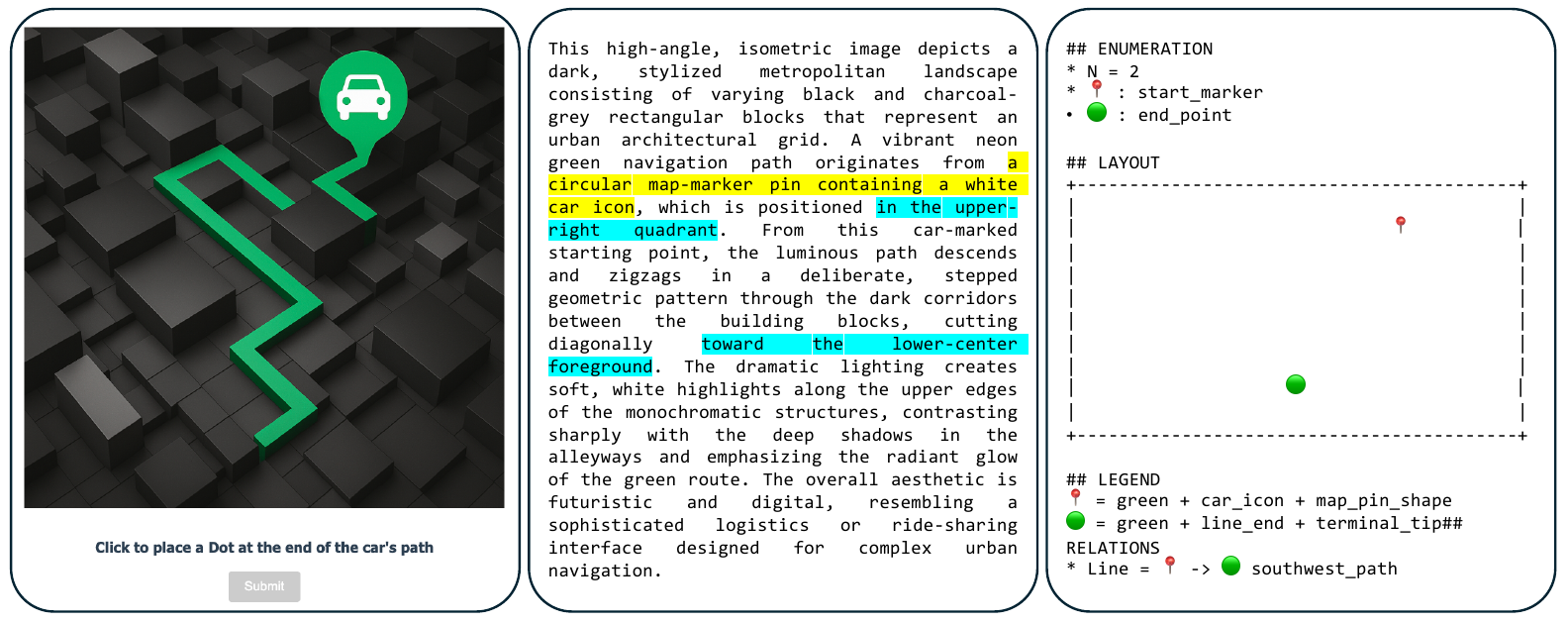}
  \caption{Visual reasoning for a \textsc{Place Dot} CAPTCHA. From left to right: the web GUI, a text-only CoT that captures \colorbox{propbg}{object attributes} and \colorbox{locbg}{coarse object locations} but may miss critical spatial relations and distances, and the proposed ASCII-box V-IR that retains the essential object count, spatial layout, attributes, and relations.}
  \label{fig:ascii_vir}
\end{figure}

%% file: sec/implementation_details.tex
\section{System Implementation Details}
\label{sec:implementation_details}

\subsection{Observation, Detection, and Local Refinement}
Each observation contains the rendered screenshot, normalized page text, and an indexed DOM snapshot. At most 120 indexed elements are forwarded to the global parser. The primary puzzle image is selected as the largest image-bearing DOM element with an area of at least 10,000 pixels squared. Object proposals within this element are produced by Grounding-DINO Tiny~\cite{liu2024grounding}. For every task family, the detector receives the same literal, class-agnostic text query, \texttt{all objects}, and uses the released text-matching threshold of 0.3. Neither the query nor the threshold is tuned per family.

Localized evidence refinement uses three pointer-centered crops. Each crop has an 8-pixel margin, a maximum side length of 512 pixels, and a window ratio of 0.2 relative to the target element extent. The local parser checks whether a candidate crop contains the evidence expected by the global plan, such as the requested symbol, face, or connection. It may adjust the coordinate or advance to a previously enumerated candidate and remains restricted to the DOM and patch indices established by the global parse. Thus, refinement corrects local grounding while preserving the object inventory committed to by V-IR.

\subsection{Tool Schema and Browser Execution}
Actions are grounded to indexed DOM elements and, when needed, normalized local coordinates. Mark is implemented as one or more grounded clicks on selectable elements; Drag additionally predicts offsets in Cascading Style Sheets (CSS) pixels, and Hold predicts a duration in milliseconds. For image-bearing targets larger than 8,000 pixels squared, the controller previews a marker at the proposed coordinate before issuing the click. The DOM index is then checked through geometric-quad visibility immediately before execution; a stale target triggers a bounded wait or re-parse.

The agent retains puzzle-scoped rather than episode-wide memory. It stores only concise facts needed by the next step, including the inferred task mode, selected region, instruction changes, and executed DOM or patch indices. Memory is reset when the backend advances to a new challenge. The implementation allows at most 20 browser interactions per puzzle, 120 seconds per model call, 300 reasoning steps per run, 10 actions per reasoning step, and 100 failures before termination.

\subsection{Transformation Admissibility}
The verification transformations are selected by manually specified family rules. Image and instruction are transformed jointly, and predicted coordinates or directions are mapped back to the original frame before signature comparison. Shifts require inverse coordinate translation; flips additionally rewrite directional terms; and color permutations rewrite the corresponding color tokens. A transformation is omitted whenever task preservation cannot be established. In particular, rotation is disabled for \textsc{Rotation Match}, and flips are disabled when direction is itself the answer. This family-specific policy evaluates action stability under declared equivalences; learning invariances automatically remains a separate direction.

%% file: sec/exp.tex
\section{Experiments}
\label{sec:experiments}

\subsection{Experimental Settings}

\noindent\textbf{Implementation.}
CaptchaAgent is implemented in approximately 2.3K lines of Python on a browser automation backend~\cite{muller2024browseruse}.
It observes page screenshots and DOM metadata and executes predicted action tuples through Playwright~\cite{microsoft2020playwright} and the Chrome DevTools Protocol.
All experiments are training-free with frozen backbones.
The consistency stage uses a bounded set of task-preserving transformations selected by puzzle family; transformations that change the requested answer are disabled.

\noindent\textbf{Datasets.}
Open CaptchaWorld~\cite{luo2025open} contains 20 puzzle families and 225 online instances with visible instructions, rendered challenge regions, and interactable DOM elements.
We also evaluate on Oedipus-data~\cite{deng2025oedipus}, an offline image-only artifact with four task groups.
Because the full Oedipus dataset is not released, we follow its repository protocol: five instances are sampled per group and each is repeated five times; textual outputs are evaluated by the protocol's LLM judge.
Together, these datasets test structured reasoning in small, discrete scenes with fixed task templates; Section~\ref{sec:limitations} discusses transfer to procedurally generated challenges and denser scenes.

\noindent\textbf{Models.}
On Open CaptchaWorld, CaptchaAgent uses GPT-4.1~\cite{gpt41}, Gemini-2.5 Pro~\cite{comanici2025gemini}, and Claude-3.7-Sonnet~\cite{anthropic_claude_sonnet_3_7_2025}.
On Oedipus-data, we use Gemini-2.0, Claude-3.7, and GPT-4o to match the reported Oedipus backbones~\cite{deng2025oedipus}.
Unless stated otherwise, decoding temperature is 1.0.
These proprietary backbones match the models available in the benchmark reports and support controller comparisons within each model family.

\noindent\textbf{Baselines and provenance.}
Open CaptchaWorld comparisons use BU-Agent, the browser-use automation baseline adopted by the benchmark~\cite{muller2024browseruse,luo2025open}.
The OpenAI-o1, GPT-4o, OpenAI-o3, Claude-3.5, and DeepSeek-V3 BU-Agent rows are transcribed from the Open CaptchaWorld report~\cite{luo2025open}; the three shared-backbone BU-Agent rows and all CaptchaAgent rows are author-run results.
Oedipus baseline rows are taken from the released Oedipus evaluation, while CaptchaAgent rows are author-run under its offline protocol.
For GPT-4.1, Gemini-2.5 Pro, and Claude-3.7-Sonnet, we ran both agents under the same benchmark protocol, making these rows the primary comparison of agent design. The transcribed rows document the broader benchmark context.

\noindent\textbf{Access and scope.}
The agent receives the rendered page, visible instructions, and browser-accessible DOM metadata available to its controller.
It may call a frozen multimodal model and a class-agnostic detector, but it has no access to CAPTCHA ground truth, server-side acceptance rules, or hidden challenge state.
Its goal is to solve each benchmark puzzle within a fixed interaction budget.
The protocol focuses on the visual and interaction layer of CAPTCHA systems; deployed defenses such as rate limits and hidden risk signals are complementary to this evaluation.

\noindent\textbf{Protocol and metrics.}
Open CaptchaWorld is evaluated online with automatic pass/fail feedback and a 20-step interaction budget per puzzle.
Oedipus-data is non-interactive and is evaluated as an image-level textual decision.
We report Pass@1 and the recorded model-call expenditure.
We report model-call expenditure separately from detector execution, browser rendering, local compute, network delay, and wall-clock time.
For failure analysis, we inspect saved observations, V-IRs, local crops, consistency traces, and final actions, assigning failures to perception-to-symbol conversion, symbolic/action reasoning, or browser execution when the trace permits.

%% file: sec/benchmark_details.tex
\subsection{Task Taxonomy and Acceptance Rules}
\label{sec:task_taxonomy}

Open CaptchaWorld covers four interaction primitives but substantially different visual and logical requirements. Table~\ref{tab:taxonomy} lists all 20 families, the action used by the environment, and the benchmark-side success condition. Set-selection tasks require an exact selected set; ordered and continuous-control tasks use stricter sequence, angle, position, or duration checks. This heterogeneity is important when interpreting aggregate Pass@1 because the same final score can conceal different perception, reasoning, and execution failures.

\begin{table*}[t]
\centering
\caption{Action-centric taxonomy and answer checking rules for the 20 Open CaptchaWorld families. ``Exact'' denotes equality with the stored answer; tolerances are instance-specific when supplied by the benchmark.}
\label{tab:taxonomy}
\scriptsize
\setlength{\tabcolsep}{3.2pt}
\begin{tabular}{llll@{\hspace{8pt}}llll}
\toprule
Family & Action & Acceptance rule & Main challenge & Family & Action & Acceptance rule & Main challenge \\
\midrule
Dice Count & Click & Exact total/option & Counting & Object Match & Click & Exact option & Cardinality \\
Geometry Click & Click & Region or 25-px tol. & Grounding & Select Animal & Mark & Exact set & Recognition \\
Rotation Match & Drag & Exact angle mod. 360 & Orientation & Coordinates & Click & Exact option & Coordinates \\
Slide Puzzle & Drag & Position, default 10 px & Translation & Path Finder & Click & Exact option & Route tracing \\
Unusual Detection & Mark & Exact set & Anomaly & Place Dot & Click & Endpoint, default 15 px & Localization \\
Image Recognition & Mark & Exact set & Recognition & Connect Icon & Click & Exact option & Relations \\
Bingo & Mark & Any winning swap pair & Board reasoning & Click Order & Click & All points in order & Sequencing \\
Image Matching & Click & Exact option & Matching & Hold Button & Hold & Timing rule & Temporal control \\
Patch Select & Mark & Exact set & Local grounding & Misleading Click & Click & Outside forbidden area & Negated grounding \\
Dart Count & Click & Exact option & Visual arithmetic & Pick Area & Click & Inside target region & Region selection \\
\bottomrule
\end{tabular}
\end{table*}

The backend checks Geometry Click inside the valid region or, for legacy point targets, within 25 pixels. Rotation Match compares the final angle modulo 360 degrees. Slide Puzzle uses the task-provided tolerance and defaults to 10 pixels. Bingo ignores swap-pair order and accepts any pair yielding a winning line. Place Dot uses an instance tolerance and defaults to 15 pixels. Click Order requires every point to match its corresponding target within tolerance and in the prescribed order. Misleading Click reverses the usual region rule: success requires staying outside the forbidden area.

\subsection{Offline Oedipus Protocol}
The Oedipus-data artifact organizes provider-specific images and answer files into four groups: \emph{Angular}, \emph{Gobang}, \emph{Space 1}, and \emph{Space 2}. We report the latter two as \emph{Space} and \emph{Space Reasoning}, respectively. Unlike Open CaptchaWorld, this protocol has no browser execution. The solver receives a static image and produces a textual decision. Gobang uses explicit analysis of a $5\times5$ board, Angular compares 6--8 directional candidate images, and the spatial groups reason over objects and relations in the image layout.

The released protocol pairs each image with a reference answer and uses an auxiliary LLM judge to compare the predicted string with the reference instruction. Semantically equivalent actions may therefore be accepted despite different wording. We follow the artifact's repeated-run setup: five instances are sampled per group and each instance is evaluated five times. Oedipus-data numbers are consequently image-level, judge-mediated accuracy and are not directly equivalent to browser-executed Open CaptchaWorld Pass@1.

%% file: sec/res.tex
\section{Results}
\label{sec:results}

\subsection{Open CaptchaWorld}
Table~\ref{tab:main} reports online Pass@1 and recorded model-call expenditure under the 20-step budget.
Across the three shared backbones, CaptchaAgent improves Pass@1 over the corresponding author-run BU-Agent configurations by 21.3--25.8 percentage points.
Gemini-2.5 Pro obtains the highest CaptchaAgent result at 46.8\%, while Claude-3.7-Sonnet gives the largest shared-backbone difference.
The matched comparisons show gains across multiple proprietary model families under the evaluated model snapshots.

\input{sec_tab/result_main}
\input{sec_fig/result_main}

The recorded model-call expenditures are similar across the matched configurations and are slightly lower for the CaptchaAgent rows.
CaptchaAgent also incurs detector, structured-parsing, local-refinement, and verification overhead; a run-aligned systems evaluation is needed to measure wall-clock efficiency.

\paragraph{Security relevance}
At 46.8\% Pass@1, CaptchaAgent solves a substantial subset of the 225 benchmark instances, showing that structured visual reasoning changes automation difficulty within these 20 families.
Section~\ref{sec:security_analysis} examines which design properties remain difficult and how the observed failures inform CAPTCHA defense.

\paragraph{Qualitative behavior}
Figure~\ref{fig:misleading_click_captcha} contrasts BU-Agent and CaptchaAgent on a misleading-click puzzle.
The baseline follows a default click prior despite describing the scene.
CaptchaAgent enumerates the layout, checks the candidate region on a local crop, and tests the action under task-preserving transformations before execution.
The trace exposes whether the final error originates in object inventory, relation inference, or grounding rather than only recording the final click.

\subsection{Oedipus-Data}
\input{sec_tab/result_oedipus}

As shown in Table~\ref{tab:addlabel}, CaptchaAgent improves average Pass@1 over the reported Oedipus solver by 12.0, 9.2, and 12.3 points for Gemini-2.0, Claude-3.7, and GPT-4o, respectively.
The largest gains occur on \textsc{Gobang} and \textsc{Space}, where object inventory and discrete relations dominate.
Gains are smaller when the answer depends on continuous attributes, consistent with the representational limits of ASCII V-IR.
The repeated, judge-mediated protocol provides a complementary image-level evaluation of the structured representation across four offline task groups.

\subsection{Ablation Study}
\input{sec_tab/ablation}

Table~\ref{tab:vir_ablation} ablates ASCII V-IR, localized refinement, and counterfactual consistency on Gemini-2.5 Pro.
Removing V-IR causes the largest observed reduction, from 46.8\% to 29.2\%.
With V-IR enabled, retaining only one of local refinement and consistency checking yields 34.8--36.8\%, compared with 46.8\% for the full pipeline.
The four measured configurations identify explicit representation as the largest contributor in this setting.
A same-budget repeated-sampling comparison remains useful for separating the effect of transformed views from naive self-consistency.

\noindent\textit{Failure modes.}
Manual trace inspection identifies three recurring bottlenecks.
First, perception-to-symbol errors omit small or occluded entities, particularly in patch-selection and unusual-detection puzzles.
Second, exact rotation, slider displacement, and other continuous attributes are poorly represented by ASCII topology.
Third, stale DOM indices or layout refreshes can invalidate an otherwise correct action.
Consistency checking detects variation across transformed views; omissions shared across all views can remain stable, and continuous parameters can remain under-specified.

%% file: sec_tab/result_main.tex
\begin{table*}[t]
  \centering
  \caption{Open CaptchaWorld results. Expenditure covers recorded model calls only, not detector, browser, network, or local-compute overhead. Legacy BU-Agent rows are reported by Open CaptchaWorld~\cite{luo2025open}; shared-backbone comparisons are author runs.}
  \label{tab:main}
  \footnotesize
  \renewcommand{\arraystretch}{0.92}
  \setlength{\tabcolsep}{16pt}
  \begin{tabular}{llcc}
  \toprule
  Agent & MLLM backbone & Pass@1 (\%) & Model-call expenditure (\$) \\
  \midrule
  \multirow{6}{*}{BU-Agent~\cite{muller2024browseruse}}
    & OpenAI-o1~\cite{jaech2024openaio1} & 5.0 & 94.6 \\
    & GPT-4o~\cite{hurst2024gpt4o} & 5.7 & 25.8 \\
    & OpenAI-o3~\cite{singh2025openaio3} & 40.0 & 66.4 \\
    & Claude-3.5-Haiku~\cite{anthropic_claude_haiku_3_5_2025} & 15.0 & 9.3 \\
    & Claude-3.5-Sonnet~\cite{anthropic_claude_sonnet_3_5_2025} & 10.0 & 21.9 \\
    & DeepSeek-V3~\cite{liu2024deepseek} & 20.0 & 7.3 \\
  \midrule
  BU-Agent & GPT-4.1~\cite{gpt41} & 25.0 & 16.7 \\
  CaptchaAgent & GPT-4.1~\cite{gpt41} & 46.3\deltaSup{+$21.3$} & 15.3 \\
  \midrule
  BU-Agent & Gemini-2.5 Pro~\cite{comanici2025gemini} & 25.0 & 18.1 \\
  CaptchaAgent & Gemini-2.5 Pro~\cite{comanici2025gemini} & 46.8\deltaSup{+$21.8$} & 16.2 \\
  \midrule
  BU-Agent & Claude-3.7-Sonnet~\cite{anthropic_claude_sonnet_3_7_2025} & 20.0 & 18.7 \\
  CaptchaAgent & Claude-3.7-Sonnet~\cite{anthropic_claude_sonnet_3_7_2025} & 45.8\deltaSup{+$25.8$} & 16.6 \\
  \bottomrule
  \end{tabular}
\end{table*}

%% file: sec_fig/result_main.tex
\begin{figure*}[t]
  \centering
  \includegraphics[width=0.98\textwidth]{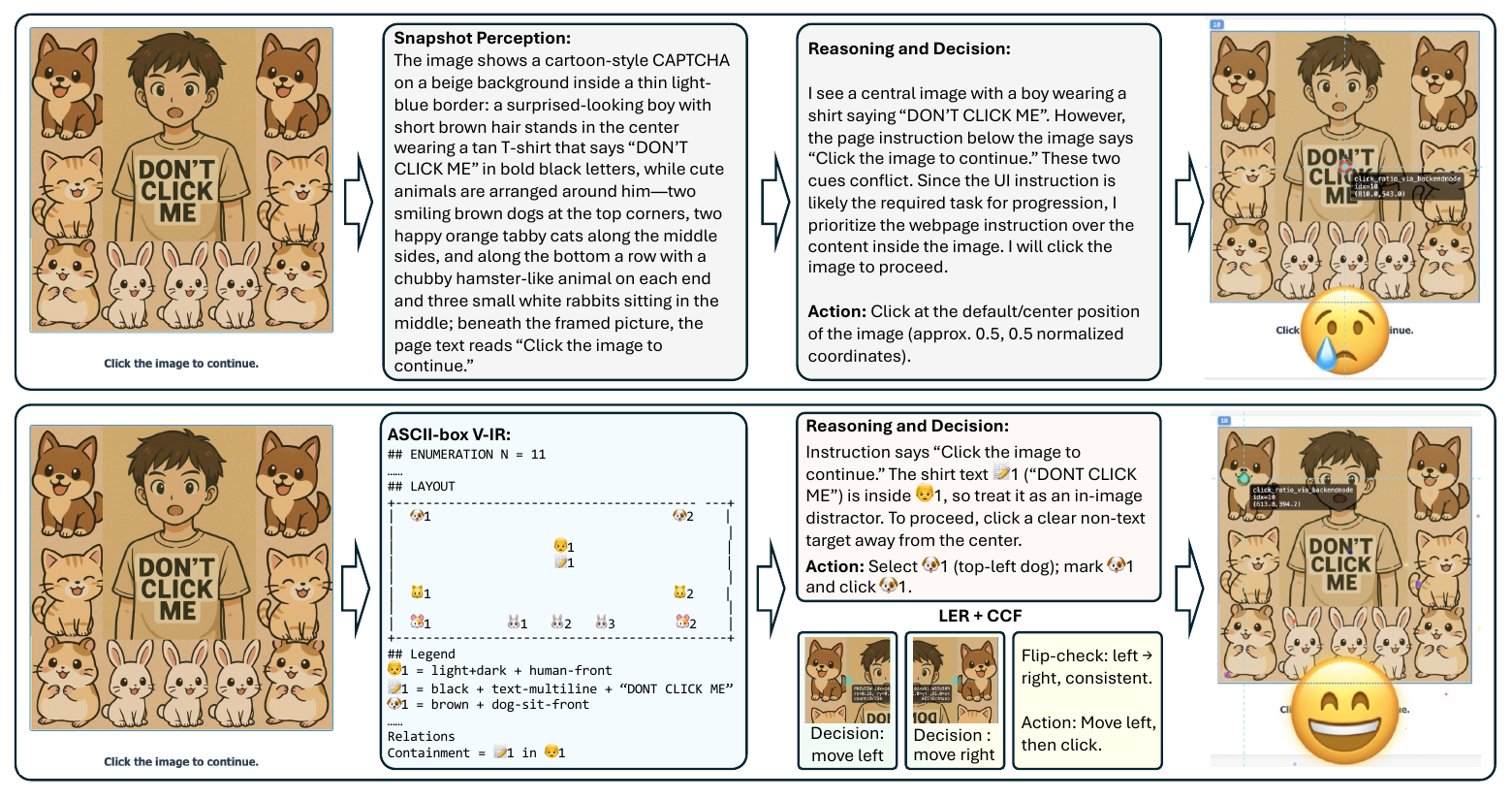}
  \caption{Qualitative comparison on misleading click CAPTCHA puzzles: the text-only BU agent~\cite{muller2024browseruse} describes the scene but clicks the default region; CaptchaAgent enumerates key layout cues and refines the click direction with localized evidence refinement (LER) and counterfactual consistency verification (CCF) to hit the correct target.}
  \label{fig:misleading_click_captcha}
\end{figure*}

%% file: sec_tab/result_oedipus.tex
\begin{table*}[t]
  \centering
  \footnotesize
  \renewcommand{\arraystretch}{0.85}
  \setlength{\tabcolsep}{8pt}
  \caption{Pass@1 (\%) on Oedipus-data~\cite{deng2025oedipus}. Oedipus rows are values reported by the released solver evaluation; CaptchaAgent rows are author runs under its offline protocol.}
    \begin{tabular}{llccccl}
    \toprule
        Agent  & Backbone & {Angular} & {Gobang} & {Space} & {Space Reasoning} & {Average} \\
    \midrule
    {\multirow{4}{*}{Oedipus~\cite{deng2025oedipus}}} & GPT-4V~\cite{gpt4v} & 37.90 & 79.70 & 71.30 & 65.00 & 63.50 \\
          & Gemini~\cite{team2023gemini} & 31.00 & 56.30 & 59.50 & 58.00 & 51.20 \\
          & miniGPT-4~\cite{zhu2023minigpt} & 7.50 & 16.90 & 18.00 & 15.10 & 14.40 \\
          & VTT~\cite{gao2021research}   &   \xmark    &    \xmark   & 55.70 & 44.50 & 25.10 \\
    \midrule
    Oedipus~\cite{deng2025oedipus} & Gemini-2.0~\cite{gemini20} & 35.10 & 79.10 & 82.70 & 62.90 & 65.00 \\
    CaptchaAgent & Gemini-2.0~\cite{gemini20} & 44.00    & 96.00    & 100.00   & 68.00    & 77.00\deltaSup{+$12.0$} \\
    \midrule
    Oedipus~\cite{deng2025oedipus} & Claude-3.7~\cite{anthropic_claude_sonnet_3_7_2025} & 42.80 & 93.40 & 80.70 & 78.20 & 73.80 \\
    CaptchaAgent & Claude-3.7~\cite{anthropic_claude_sonnet_3_7_2025} & 48.00    & 100.00   & 96.00    & 88.00    & 83.00\deltaSup{+$9.2$} \\
    \midrule
    Oedipus~\cite{deng2025oedipus} & GPT-4o~\cite{hurst2024gpt4o} & 37.20 & 78.80 & 70.90 & 63.70 & 62.70 \\
    CaptchaAgent & GPT-4o~\cite{hurst2024gpt4o} & 44.00    & 96.00    & 92.00    & 68.00    & 75.00\deltaSup{+$12.3$} \\
    \bottomrule
    \end{tabular}%
  \label{tab:addlabel}%
\end{table*}%

%% file: sec_tab/ablation.tex
\begin{table*}[t]
\centering
  \footnotesize
\renewcommand{\arraystretch}{0.85}
  \setlength{\tabcolsep}{4pt}
  \caption{Ablation on Open CaptchaWorld~\cite{luo2025open} using Gemini-2.5 Pro~\cite{comanici2025gemini}: impact of ASCII-based V-IR, localized refinement, and counterfactual consistency on Pass@1.}
\label{tab:vir_ablation}
\begin{tabular}{lccc}
\toprule
ASCII-based V-IR & Localized Refinement & Counterfactual Consistency & Pass@1 [\%] \\
\midrule
      & \checkmark & \checkmark & 29.2 \\
\checkmark &         & \checkmark & 36.8 \\
\checkmark & \checkmark &         & 34.8 \\
\checkmark & \checkmark & \checkmark & 46.8 \\
\bottomrule
\end{tabular}
\end{table*}

%% file: sec/qualitative_details.tex
\subsection{Family-Level and Qualitative Analysis}
\label{sec:qualitative_details}

The traces indicate that performance is uneven across task families. Interaction-heavy and trap-like tasks, including \textsc{Hold Button}, \textsc{Misleading Click}, and \textsc{Bingo}, benefit from explicit action state and re-observation. Tasks dominated by exact counting, endpoint localization, or rigid ordered execution remain difficult; recurrent failures occur on \textsc{Dice Count}, \textsc{Place Dot}, and \textsc{Click Order}. We use these traces for qualitative diagnosis, while the aggregate tables remain the basis for quantitative comparison.

\input{sec_fig/connect_icon_trace}

Figure~\ref{fig:connect_icon_trace} compares a \textsc{Connect Icon} reference with two candidate regions and their ASCII V-IRs. The example shows why the three modules address different failure locations. V-IR exposes the icon inventory and spatial relations, local crops revisit fine evidence around the candidates, and transformed views test whether the resulting action remains stable. If every view omits the same small icon, consistency voting can still preserve a stable but incorrect plan. Nevertheless, the shared omission remains visible in the structured trace.

%% file: sec_fig/connect_icon_trace.tex
\begin{figure*}[t]
    \centering
    \includegraphics[width=0.94\textwidth]{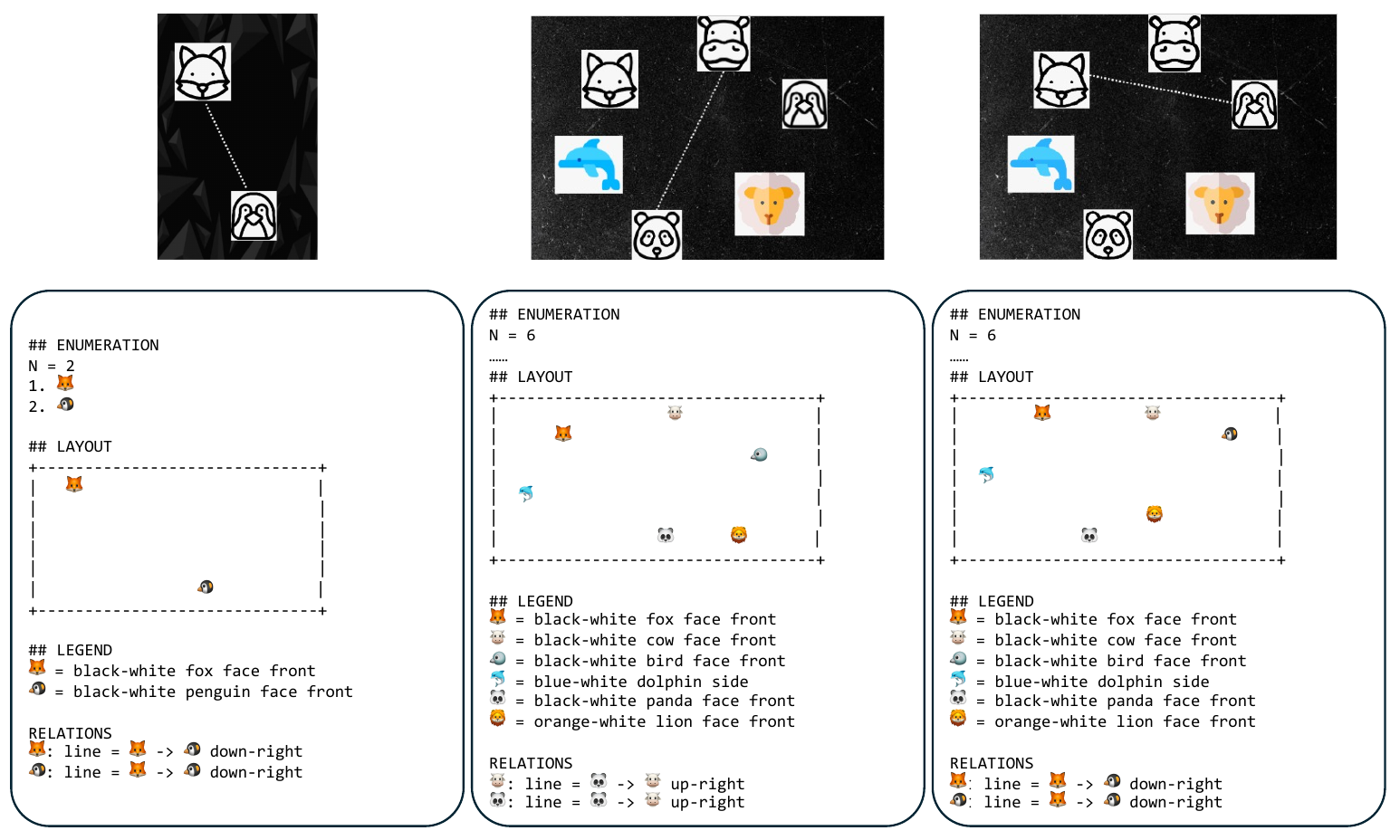}
    \caption{Structured trace for a \textsc{Connect Icon} instance. The reference region and two candidate regions are paired with their ASCII V-IRs, which expose the icon inventories, spatial layouts, and directed connections before action selection.}
    \label{fig:connect_icon_trace}
\end{figure*}

%% file: sec/security_analysis.tex
\section{Security Implications for CAPTCHA Design}
\label{sec:security_analysis}

CaptchaAgent's intermediate traces expose where a visual challenge continues to impede automation. A successful attempt requires a usable page observation, a sufficiently complete visual state, a valid decision, and correct execution before the challenge changes state. Missing targets indicate observation limits; incomplete inventories or relations indicate representation limits; unstable or consistently wrong plans indicate decision limits; and timing or page refreshes indicate execution limits. Because the shared-backbone comparisons hold the VLM family fixed, their gains show that controller structure itself changes solving success on the evaluated challenges.

The traces suggest that adding more discrete objects or pairwise relations is unlikely to provide a durable barrier when the full state can be externalized into a compact entity--relation representation. Procedural generators can increase automation difficulty by varying the underlying relation graph and checking rule, not only visual appearance. Precise orientation, trajectory, timing, endpoint geometry, and ordered control remain harder for the current controller because they require metric or sequential state that the ASCII V-IR represents only coarsely. These observations concern the visual-reasoning layer of CAPTCHA security. Behavioral bot classifiers and webpage-security detectors can provide evidence outside the puzzle image~\cite{cabri2022quantum,purwanto2022phishsim}, while rate limits, replay resistance, risk-based escalation, and server-side state changes provide further complementary protection.

Automation resistance must also be balanced against usability and accessibility. Security-assistive interfaces can improve human attention without simply increasing perceptual or motor burden~\cite{huang2022advert}. Visual density and motor precision may increase machine difficulty while disproportionately burdening legitimate users. Practical deployments should therefore evaluate human completion cost together with automated success, rotate challenge generators, and retire families whose security margin deteriorates under contemporary agents.

%% file: sec/limitations.tex
\section{Limitations, Reproducibility, and Safety}
\label{sec:limitations}

\noindent\textit{Scope and representation.}
CaptchaAgent is designed for low-entropy CAPTCHA scenes with countable objects, bounded actions, and explicit success checks. Open CaptchaWorld contains 20 fixed families, and Oedipus-data contains four small offline groups; tests on procedural generators and deployed challenge families are needed to establish transfer. ASCII V-IR represents discrete spatial and topological relations effectively but loses continuous geometry, subtle gradients, uncertainty, and fine rotation angles. Exhaustive pairwise relation enumeration also scales quadratically with object count, so denser scenes would require sparse relation selection and an explicit metric representation. Missed detector boxes or incorrect relations can propagate to the final action, and consistency checking cannot correct an omission repeated across all transformed views.

\noindent\textit{Evaluation and reproducibility.}
The online study contains 225 instances and reports single-run point estimates. Paired confidence intervals, significance tests, and larger procedurally generated evaluations would strengthen the empirical evidence. The evaluated proprietary backbones are model snapshots, and reproduction depends on continued API access; open-weight and newer-model experiments would better separate controller gains from changing backbone capability. The paper and supplementary material specify prompts, tool mappings, detector settings, action budgets, and benchmark protocols. Recorded model-call expenditure excludes detector time, browser rendering, network delay, and local compute, so a run-aligned systems study should additionally report wall-clock latency, call and token counts, and hardware utilization.

\noindent\textit{Responsible use.}
The experiments use local or research benchmark instances to study CAPTCHA solving and defense. They do not target production services, credentials, or hidden risk-control systems. A responsible release should separate benchmark analysis components from service-specific operational adapters and coordinate newly identified weaknesses with benchmark or service maintainers before disclosure.

%% file: sec/concl.tex
\section{Conclusion}
We presented CaptchaAgent, a structured visual reasoning framework for interactive CAPTCHA solving. Its V-IR and ASCII box diagram commit the parser to an explicit object inventory, localized refinement revisits ambiguous visual evidence, and task-preserving transformations expose unstable action signatures before execution. Across matched proprietary backbones, this structure improves Pass@1 over the evaluated browser-use baseline, and the component study shows complementary gains from explicit representation, local refinement, and consistency checking. The intermediate traces further show that discrete relational tasks and interaction traps are amenable to structured solving, whereas exact counting, continuous geometry, ordered control, and browser-state changes remain the main failure sources. These findings connect method-level improvements to CAPTCHA design and motivate challenge generators that combine relational reasoning with precise metric and sequential requirements.

%% file: sec/supp.tex
\noindent\textbf{Overview.}
This supplementary document adds implementation-level details that complement the main paper. Section~\ref{sec:taxonomy} summarizes the 20 CAPTCHA families in OpenCaptchaWorld~\cite{luo2025open}. Section~\ref{sec:prompts} presents the prompt templates used by CaptchaAgent, including the system prompt, the global visual reasoning prompt, the ASCII-based intermediate-representation prompt, and the localized refinement prompt. Section~\ref{sec:grounding} details webpage observation, puzzle-image selection, patch grounding with Grounding-DINO~\cite{liu2024grounding}, and action execution. Section~\ref{sec:checking} lists the benchmark-side answer checking rules. Section~\ref{sec:hyper} records the hyperparameters and protocol settings used in CaptchaAgent and in the experiments described in the main paper. Section~\ref{sec:qual} provides the complete webpage context and local visual evidence for a representative \textsc{Connect Icon} case. We also include a brief discussion of the current family-level limitations of the method.

\section{CAPTCHA Task Taxonomy}
\label{sec:taxonomy}

Open CaptchaWorld~\cite{luo2025open} contains 20 CAPTCHA families, summarized in Table~\ref{tab:taxonomy}. The action column uses the primitive interaction families \emph{Click}, \emph{Mark}, \emph{Drag}, and \emph{Hold}; none of the evaluated families requires free-form typing or scrolling as its core action.

\begin{table*}[t]
\centering
\caption{Compact taxonomy of the 20 Open CaptchaWorld CAPTCHA families.}
\label{tab:taxonomy}
\footnotesize
\setlength{\tabcolsep}{5pt}
\begin{tabular}{p{0.17\textwidth}p{0.08\textwidth}p{0.27\textwidth}p{0.37\textwidth}}
\toprule
Type & Action & Success criterion & Main challenge \\
\midrule
Dice Count & Click & exact option or total & count top faces correctly \\
Geometry Click & Click & inside valid region or tolerance & geometric grounding \\
Rotation Match & Drag & exact angle modulo 360 & orientation alignment \\
Slide Puzzle & Drag & final position within tolerance & translation estimation \\
Unusual Detection & Mark & exact selected set & anomaly detection \\
Image Recognition & Mark & exact selected set & category recognition \\
Bingo & Mark & any valid winning swap pair & counterfactual board reasoning \\
Image Matching & Click & exact option & reference matching \\
Patch Select & Mark & exact selected set & local object grounding \\
Dart Count & Click & exact option & visual arithmetic \\
Object Match & Click & exact option & cardinality matching \\
Select Animal & Mark & exact selected set & fine-grained recognition \\
Coordinates & Click & exact option & coordinate interpretation \\
Path Finder & Click & exact option & route tracing \\
Place Dot & Click & click within tolerance & endpoint localization \\
Connect Icon & Click & exact option & relational matching \\
Click Order & Click & all points correct and ordered & identity plus order \\
Hold Button & Hold & duration satisfies rule & temporal control \\
Misleading Click & Click & remain outside forbidden area & negated grounding \\
Pick Area & Click & inside correct region & region selection \\
\bottomrule
\end{tabular}
\end{table*}

\section{Prompt and Protocol Details}
\label{sec:prompts}

This section presents the prompt templates used by CaptchaAgent. For readability, we lightly normalize line breaks and remove implementation-only wrappers, while preserving the actual instruction content and decision constraints.

\subsection{System prompt}
The system prompt constrains the agent to one tool invocation per step and restricts the available action space to the browser-grounded tools used in our method:

\begin{lstlisting}
CAPTCHA agent. One action per step.

OUTPUT MUST be valid JSON:
{"action":[{...}]}

AllowedActions (MUST choose exactly one):
- ascii_scene_ir
- reason_step
- preview_index_clickpoint
- click_ratio_via_backendnode
- hold_via_backendnode
- drag_by_offset

NeverDone:
- DO NOT use "done".
- DO NOT output tool calls as plain text.
- If unsure, call reason_step.
\end{lstlisting}

The first two entries are reasoning tools rather than browser actions. The executable tools map to the action families used elsewhere in the paper as follows: \texttt{preview\_index\_clickpoint} and \texttt{click\_ratio\_via\_backendnode} implement Click/Mark, \texttt{hold\_via\_backendnode} implements Hold, and \texttt{drag\_by\_offset} implements Drag. Scroll, typing, waiting, and navigation are browser-agent capabilities shown in the general framework, but they are not core actions for the Open CaptchaWorld tasks evaluated here.

\subsection{Global visual reasoning prompt}
The global visual reasoning stage consumes the current webpage screenshot, webpage text, and indexed DOM elements, and returns a structured next-step decision. The prompt used by CaptchaAgent is shown below.

\begin{lstlisting}
You are a VISION PARSER for CAPTCHA puzzles. Return ONLY valid JSON.

BUDGET:
- total steps remaining
- total puzzles remaining

INPUTS:
- current URL
- puzzle signature
- normalized page text
- indexed webpage elements (index, text, bounding box, tag)
- page screenshot

OUTPUT JSON MUST include:
- puzzle_type
- plan
- candidates
- notes

Puzzle types:
slider, geometry_click, pick_area, click_order,
rotation_match, counting, match_this, unknown

Plan fields:
- action: click | drag | hold | multi | none
- target element index
- local normalized x/y in [0,1]
- drag offsets in CSS pixels
- optional submit element index

MATCH_THIS rule:
1) describe the left reference target;
2) select the most similar item on the right;
3) assign stable ids to the left target and to each right-side option.

V-IR requirements:
- enumerate visible targets explicitly;
- keep short attributes for name, size, direction, shape, color,
  material, and anomaly;
- record pairwise relations such as left-of, inside, overlap,
  alignment, and connection;
- if unsure, output puzzle_type="unknown" and action="none".
\end{lstlisting}

The structured decision produced by this stage is the executable plan $\pi_t$ introduced in the main paper. The parser also returns a ranked candidate list for subsequent localized refinement.

\subsection{ASCII intermediate-representation prompt}
The ASCII V-IR prompt is responsible for the inspectable box-diagram representation described in Section 3 of the main paper. The prompt used by our implementation is reproduced below in a Unicode-safe form for \LaTeX{} compilation.

\begin{lstlisting}
TASK: Output a Markdown ASCII box diagram using ONLY "+", "-",
"|", and spaces to represent the relative spatial positions of all
TRUE targets in the cropped puzzle image.

HARD CONSTRAINTS
1) Inside the box: only minimal markers. No natural language.
2) Repeated markers must receive unique numeric suffixes.
3) Enumerate all visible targets before drawing the diagram.
4) Output the exact locked target count N.
5) The box and legend must contain exactly N targets.
6) Do not invent or complete occluded objects.
7) Output one legend line per target.
8) Enumerate all special relations: contain, overlap, under/over,
   link, touch. If none exist, output "None special relations".
9) Ignore background; describe targets only.
10) Text targets must be preserved as targets.
11) Relative positions must reflect the true viewpoint.
12) For dice: read only the physically upward-facing face.

OUTPUT FORMAT
A) LockedTargets: N=...
B) ASCII box diagram
C) Legend
D) Relations
\end{lstlisting}

This prompt operationalizes the three constraints highlighted in the main paper: exact object-count commitment, symbol-only spatial markup inside the box, and exhaustive relation reporting outside the box.

\subsection{Localized evidence refinement prompt}
Localized evidence refinement re-checks the current decision around the intended interaction locus using up to three higher-resolution local crops. The prompt is:

\begin{lstlisting}
Return ONLY JSON.

You are a planner that refines a global visual representation using
local patches.
- Each patch is a local view at or near the planned click position.
- Check whether the patch matches the expected local evidence for a
  correct click (e.g., correct letter, object, face, or symbol).
- If it does not match, revise the local position or move to the next
  candidate.
- Do not invent new target indices.
- The final plan must stay compatible with the global action schema.

Inputs:
- current URL
- puzzle signature
- normalized page text
- global visual representation
- metadata for the candidate patches
- the patch images themselves
\end{lstlisting}

\subsection{Counterfactual consistency protocol}
Counterfactual consistency verification follows the main paper: the current plan is re-parsed under a broader set of semantics-preserving image transformations, including rotation, flip, shift, and color permutation, and agreement is checked on the action-relevant signature, i.e., the action type, the target element, the local click point, drag offsets, and any optional submit target. When transformed views disagree, CaptchaAgent resolves the discrepancy through majority voting over the transformed decision signatures rather than relying on model-reported confidence.

\section{Webpage Processing, Grounding, and Execution}
\label{sec:grounding}

\subsection{Observation and primary puzzle image}
As described in the main paper, each observation is $o_t=\{I_t,x_t,e_t\}$, where $I_t$ is the current page screenshot, $x_t$ is the on-page text, and $e_t$ is the indexed DOM snapshot. The implementation forwards at most 120 indexed elements to the global reasoning stage. The primary puzzle image $J_t$ is then selected as the largest image-bearing DOM element, following the formal definition in the main paper.

\subsection{Patch grounding with Grounding-DINO}
To obtain object-level references inside $J_t$, we use Grounding-DINO Tiny~\cite{liu2024grounding}. The detector is queried with the same class-agnostic text prompt, \texttt{all objects}, for all 20 Open CaptchaWorld families; no task-specific object vocabulary is supplied. We use the released text-matching threshold of 0.3 without per-category tuning. Detected boxes are represented in the coordinate frame of $J_t$, allowing the same reference to be mapped back to normalized element-local coordinates for browser actions.

\subsection{Localized crop parameters}
Localized evidence refinement uses pointer-centered local crops with the following values:
\begin{itemize}
    \item number of local crops: 3;
    \item crop margin: 8 pixels;
    \item maximum crop side length: 512 pixels;
    \item local crop window ratio: 0.2 relative to the target element extent.
\end{itemize}
These settings determine the effective visual resolution used by LER near the predicted click point.

\subsection{Action grounding and execution}
The execution layer supports four primitive interaction families: Click, Mark, Drag, and Hold. Mark is implemented as one or more grounded click operations on selectable elements, while Drag and Hold use their dedicated backend tools. Each action is grounded to an indexed DOM element and, when required, a normalized local coordinate within that element. Drag actions additionally predict horizontal and vertical offsets in CSS pixels. Hold actions additionally predict a duration. For large image-bearing targets, the agent uses a preview-before-click protocol: it first renders a preview marker at the proposed local coordinate and executes the actual click only after the marker location is visually confirmed.

\subsection{Puzzle-scoped memory}
CaptchaAgent maintains short-term memory at the puzzle level rather than the episode level. The memory stores only concise step-relevant facts, such as the inferred task mode, the chosen target region, and the current local interaction parameters. This memory is carried across steps for the same puzzle and reset when the environment advances to the next challenge.

\section{Benchmark Answer Checking Rules}
\label{sec:checking}

The Open CaptchaWorld backend~\cite{luo2025open} evaluates each family with task-specific rules. We summarize the relevant acceptance criteria below.
\begin{itemize}
    \item Dice Count requires exact equality with the stored count or sum.
    \item Geometry Click accepts either a click inside the target region or, for legacy instances, a click within 25 pixels of the target point.
    \item Rotation Match requires the final angle to match exactly modulo 360 degrees.
    \item Slide Puzzle requires the final position to lie within the task-specific tolerance, which defaults to 10 pixels.
    \item Unusual Detection, Image Recognition, Patch Select, and Select Animal require exact set match.
    \item Bingo accepts any swap pair that produces a valid winning line; pair order is ignored.
    \item Image Matching, Dart Count, Object Match, Coordinates, Path Finder, and Connect Icon require the exact option identity.
    \item Place Dot accepts a click within the task-specific tolerance around the target endpoint; the default tolerance is 15 pixels when not otherwise specified.
    \item Click Order requires every clicked point to match the corresponding target within tolerance and to appear in the correct order.
    \item Hold Button requires the hold duration to satisfy the puzzle-specific timing rule.
    \item Misleading Click is correct only if the click remains outside the forbidden area.
    \item Pick Area requires the click to lie inside the correct region.
\end{itemize}

\subsection{Implementation details for Oedipus-data}
In addition to Open CaptchaWorld~\cite{luo2025open}, we evaluate CaptchaAgent on Oedipus-data~\cite{deng2025oedipus} under the offline image-only protocol described in the main paper. The Oedipus solver artifact organizes the benchmark as a set of provider-specific image folders together with task-specific answer files. In our paper, we report the four task groups \emph{Angular}, \emph{Gobang}, \emph{Space}, and \emph{Space Reasoning}; in the released Oedipus artifact, the two spatial groups are organized as \emph{Space 1} and \emph{Space 2}, which we align with \emph{Space} and \emph{Space Reasoning}, respectively.

The evaluation pipeline is structurally similar to our Open CaptchaWorld setup, but without browser interaction. For each offline instance, the solver loads the challenge image, issues task-specific multimodal prompts, and returns a textual solution rather than a DOM-grounded action. The Oedipus artifact uses a unified multi-model solver that dispatches to task-specialized reasoning routines for the different CAPTCHA families. In particular, Gobang is handled by explicit grid analysis over a 5$\times$5 board; Angular is handled by multi-image directional analysis over 6--8 candidate images; and the two spatial families are handled by object- and relation-centric reasoning over the static image layout. For some task families, the artifact further accelerates inference through parallel sub-analyses over rows, columns, or candidate images.

Answer checking on Oedipus-data is also text-based rather than environment-based. The artifact pairs each benchmark image with a reference answer file and uses an auxiliary LLM judge to compare the predicted solution string against the ground-truth instruction. The judge treats two answers as correct when they describe the same effective action even if their wording differs. This offline evaluation protocol is the reason why our Oedipus-data experiments are reported as image-level accuracy rather than browser-executed success rate.

Consistent with the protocol described in the main paper, we sample 5 instances per task type and repeat each instance 5 times. The released Oedipus reproduction script is similarly organized around repeated multi-model runs over a fixed set of benchmark categories, and the artifact logs both completion statistics and LLM-judged accuracy for the resulting text answers.

\section{Hyperparameters and Protocol Settings}
\label{sec:hyper}

\paragraph{Experiment protocol}
The main paper evaluates CaptchaAgent on Open CaptchaWorld~\cite{luo2025open} with a 20-step interaction budget per puzzle and on Oedipus-data~\cite{deng2025oedipus} with 5 sampled instances per type and 5 repeated runs per instance. All experiments are training-free and use frozen multimodal backbones. Unless otherwise stated in the main paper, decoding temperature is set to 1.0.
\paragraph{Action and budget settings}
We use the following control limits:
\begin{itemize}
    \item maximum reasoning steps per run: 300;
    \item maximum actions per step: 10;
    \item maximum failures before termination: 100;
    \item maximum puzzle count per run: 225;
    \item model timeout per call: 120 seconds.
\end{itemize}
\paragraph{Grounding and refinement settings}
We use the following grounding hyperparameters:
\begin{itemize}
    \item maximum number of DOM elements forwarded to the parser: 120;
    \item minimum area threshold for selecting the primary puzzle element: 10{,}000 pixels squared;
    \item large-element threshold for preview-before-click: 8{,}000 pixels squared;
    \item detector text threshold: 0.3;
    \item number of local refinement crops: 3;
    \item local crop margin: 8 pixels;
    \item maximum local crop side length: 512 pixels;
    \item local crop window ratio: 0.2.
\end{itemize}
\paragraph{Implementation stack}
CaptchaAgent is implemented on a browser automation backend following the setup described in the main paper, with browser interaction executed through Playwright~\cite{microsoft2020playwright} and the Chrome DevTools Protocol. The browser-agent substrate follows browser-use~\cite{muller2024browseruse}.

\section{Current Limitations by CAPTCHA Family}

The present system remains uneven across CAPTCHA families. Its strongest gains appear on interaction-heavy or trap-like puzzles such as \textsc{Hold Button}, \textsc{Misleading Click}, and \textsc{Bingo}, where explicit intermediate structure, localized evidence refinement, and stepwise verification substantially improve action selection. However, performance is still limited on families whose difficulty is dominated by exact counting, endpoint localization, or rigid ordered execution. In particular, \textsc{Dice Count}, \textsc{Place Dot}, and \textsc{Click Order} remain failure-prone in the current system, with success rates still close to zero in practical runs. This gap indicates that while CaptchaAgent already improves structured visual grounding, it still needs stronger mechanisms for exact counting, fine-grained spatial endpoint inference, and sequence-level control.

\section{Qualitative Examples}
\label{sec:qual}

\begin{figure}[t]
    \centering
    \includegraphics[width=0.92\linewidth]{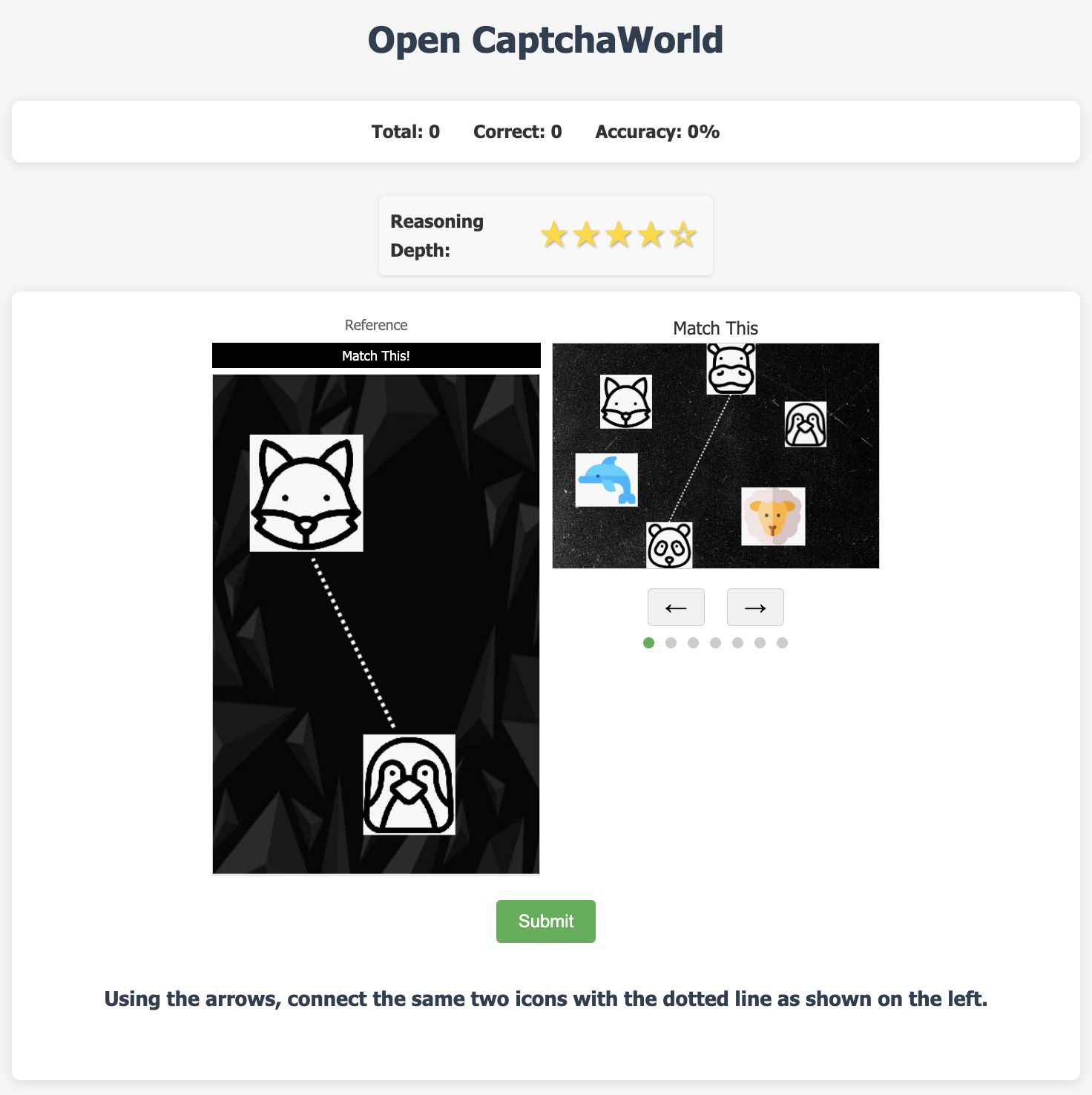}
    \caption{A full-page screenshot of a \textsc{Connect Icon} puzzle in Open CaptchaWorld. The webpage contains the textual instruction, the left-side reference icon set, and the right-side candidate regions from which CaptchaAgent must identify the matching option.}
    \label{fig:ocw_fullpage_example}
\end{figure}

\begin{figure}[t]
    \centering
    \begin{minipage}{0.19\linewidth}
        \centering
        \includegraphics[width=\linewidth]{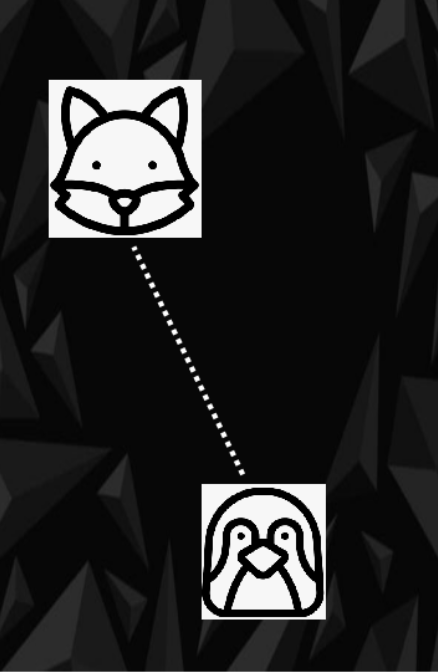}
        \vspace{1mm}

        {\scriptsize (a) Reference}
    \end{minipage}\hfill
    \begin{minipage}{0.37\linewidth}
        \centering
        \includegraphics[width=\linewidth]{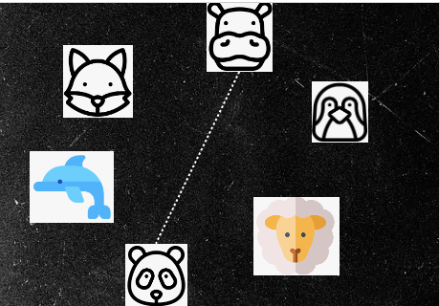}
        \vspace{1mm}

        {\scriptsize (b) Candidate region 1}
    \end{minipage}\hfill
    \begin{minipage}{0.37\linewidth}
        \centering
        \includegraphics[width=\linewidth]{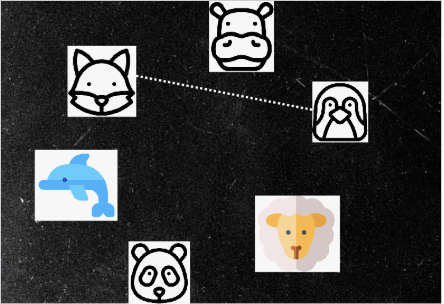}
        \vspace{1mm}

        {\scriptsize (c) Candidate region 2}
    \end{minipage}
    \caption{Local crops from the same \textsc{Connect Icon} instance. The left crop is the reference icon set to be matched. The two right crops are candidate regions containing both matching icons and distractor icons, making the task depend on explicit icon inventory rather than coarse global similarity alone.}
    \label{fig:connect_icon_crops}
\end{figure}

We provide additional context for the representative \textsc{Connect Icon} case discussed in the main manuscript. Figure~\ref{fig:ocw_fullpage_example} shows the complete webpage, and Figure~\ref{fig:connect_icon_crops} isolates the reference icon set together with two candidate regions. These views complement the structured trace in the main manuscript by exposing the original page context and the local visual evidence supplied to the parser.

